\documentclass[letterpaper, 10pt, conference]{ieeeconf}

\usepackage[T1]{fontenc}

\IEEEoverridecommandlockouts
\makeatletter
\let\NAT@parse\undefined
\makeatother

\usepackage{subcaption}
\usepackage{graphicx}
\usepackage{amsmath,amssymb}
\usepackage[colorlinks=true]{hyperref}
\usepackage{flushend}
\usepackage[acronym,nomain]{glossaries}

\newacronym{uav}{UAV}{Unmanned Aerial Vehicle}
\newacronym{ga}{GA}{Genetic Algorithm}
\newacronym{pso}{PSO}{Particle Swarm Optimisation}
\newacronym{ga-pso}{GA-PSO}{Genetic Algorithm-Particle Swarm Optimisation}
\newacronym{pmx}{PMX}{Partially Matched Crossover}
\newacronym{udp}{UDP}{User Datagram Protocol}
\newacronym{anova}{ANOVA}{Analysis of Variance}
\newacronym{ci}{CI}{Confidence Interval}
\newacronym{aco}{ACO}{Ant Colony Optimisation}
\newacronym{nsgaii}{NSGA-II}{Non-dominated Sorting Genetic Algorithm II}
\newacronym{moeadde}{MOEA/D-DE}{Multiobjective Evolutionary Algorithm Based on Decomposition with Differential Evolution}
\glsdisablehyper
\makeglossaries

\graphicspath{{.}{Figures/}{Styles/}}

\makeatletter
\renewcommand{\abstract}{\normalfont%
    \if@twocolumn%
      \@IEEEabskeysecsize\bfseries\textit{Abstract.}\ %
    \else%
      \begin{center}\vspace{-1.78ex}\@IEEEabskeysecsize\textbf{Abstract}\end{center}\quotation\@IEEEabskeysecsize%
    \fi\@IEEEgobbleleadPARNLSP}
\makeatother

\title{\LARGE \bf TriSAR: Task Coordination and Collision Avoidance for Aerial Robot Teams in Disaster Response}

\author{\authorblockN{Aditya Anil Kapile\authorrefmark{1}, Pedro Machado\authorrefmark{1}, Isibor Kennedy Ihianle\authorrefmark{1}}
\authorblockA{\authorrefmark{1}School of Science and Technology, Nottingham Trent University, Nottingham, United Kingdom}
}

\begin{document}

\maketitle
\thispagestyle{empty}
\pagestyle{empty}

\begin{abstract}
Multi-Unmanned Aerial Vehicle (UAV) disaster-response systems require coordinated task assignment and local trajectory control, yet the individual and combined contributions of these coordination layers to mission efficiency and operational safety remain insufficiently characterised under controlled experimental conditions. TriSAR is evaluated as a five-UAV coordination system operating in a physics-based Gazebo simulation of an earthquake-damaged urban environment. A $2\times2$ factorial design compares two task-allocation strategies (Genetic Algorithm and greedy fitness-based allocation) with reactive collision avoidance either enabled or disabled. Each of the four configurations was evaluated over 30 stochastic episodes drawn from a single consistent execution batch, within a common scenario of five UAVs and eight targets. The results reveal an allocation-dependent safety pattern. Under greedy allocation, enabling repulsion eliminated recorded collision-threshold violations relative to the disabled condition, a difference confirmed by a Mann-Whitney test ($U=885$, $p=4.03\times10^{-12}$, rank-biserial $r=0.97$). Under GA allocation, the same protective effect was also confirmed (Mann-Whitney $U=675$, $p=1.26\times10^{-5}$, rank-biserial $r=0.50$), though with a smaller effect size than under greedy allocation. For the continuous mission-efficiency metrics, GA-based allocation showed no statistically detectable advantage over greedy allocation when repulsion was enabled, but a significant advantage on all three efficiency metrics (steps, path length, energy) when repulsion was disabled (Welch's $t$-tests, $|g|$ between 0.92 and 1.76). The findings therefore show that reactive repulsion provides a substantial, allocation-dependent safety benefit, while the additional computational complexity of GA-based task allocation yields a detectable mission-efficiency benefit only in the specific condition where repulsion is disabled.
\end{abstract}

\vspace{0.5em}

\section{Introduction}

Murphy \cite{murphy2014} highlights how earthquakes, structural collapse, and post-disaster fires can restrict ground access for human first responders at precisely the stage when rapid survivor localisation is most critical. \Glspl{uav} provide an alternative means of accessing hazardous or obstructed areas, while coordinated aerial robot teams can distribute search, structural assessment, and communication-relay tasks across a wider operational area than a single platform, a capability demonstrated in deployed multi-UAV search-and-rescue systems \cite{scherer2015}. Alqefari and Menai \cite{alqefari2025} identify dynamic multi-UAV task assignment as a central challenge in such systems, particularly when mission priorities, resource constraints, and environmental conditions vary during operation. Effective disaster-response coordination therefore requires two closely related decisions: which aerial robot should be assigned to each mission target, and how each robot should navigate towards its assignment while maintaining safe separation from neighbouring agents and obstacles.

Yan et al. \cite{yan2021pso} and Xiong and Zhang \cite{xiong2025greapso} demonstrate the potential of hybrid optimisation approaches that combine task allocation with trajectory or route planning for multi-UAV systems. However, evaluations of such architectures commonly focus on aggregate performance against alternative optimisation methods, making it difficult to determine how individual coordination components contribute to observed mission outcomes. The research gap addressed in TriSAR is therefore not the combination of \gls{ga} and \gls{pso} itself, but the limited controlled evidence concerning the contribution of each coordination layer. In particular, it remains unclear whether the additional search complexity introduced by GA-based allocation provides a measurable benefit for small aerial robot teams, whether reactive repulsion improves operational safety, and whether the observed safety contribution of reactive control depends on the task-allocation strategy.

The study addresses three research questions. RQ1: Does GA-based task-priority optimisation improve mission efficiency relative to greedy fitness-based allocation in the tested multi-UAV disaster-response scenario? RQ2: Does reactive repulsion reduce the incidence and number of collision-threshold violations? RQ3: Does the observed safety effect of reactive repulsion depend on the task-allocation strategy? RQ1 and RQ2 are associated with directional hypotheses defined before analysis. H1: GA-based allocation is expected to reduce mission cost relative to greedy fitness-based allocation, as reflected by execution steps, fleet path length, and simulated energy consumption. H2: enabling reactive repulsion is expected to reduce the incidence and number of collision-threshold violations relative to configurations in which repulsion is disabled. RQ3 remains exploratory because the allocation-dependent safety pattern was identified during analysis of the experimental results in Section VI, rather than being specified as a directional hypothesis in advance.

The article makes three contributions. First, it provides an integrated \gls{ga-pso} coordination architecture combining a permutation-encoded GA task allocator with a PSO-based trajectory controller and reactive repulsion within a physics-based Gazebo Sim environment. Second, a controlled $2\times2$ factorial ablation separates task-allocation strategy from reactive collision avoidance, allowing the individual and combined effects of the two coordination layers to be examined rather than evaluating only the complete architecture. Third, the study reports a conditional pattern rather than a uniform one: under the investigated configuration of five aerial robots and eight mission targets, GA-based allocation shows no statistically detectable mission-efficiency advantage over greedy allocation when reactive repulsion is enabled, but a significant advantage on all three efficiency metrics when repulsion is disabled, while reactive repulsion itself provides a safety benefit that is statistically confirmed under both allocators, with a substantially larger effect size under greedy allocation than under GA allocation.

\section{Related Work}

Swarm-intelligence reviews \cite{nguyen2024,duan2023} and multi-robot task-allocation taxonomies \cite{gerkey2004,chakraa2023} together establish that most coordinated robot-team systems combine a global assignment mechanism with local, reactive rules; TriSAR investigates precisely this two-layer structure. Gerkey and Matari\'{c}'s taxonomy \cite{gerkey2004} classifies TriSAR's own allocator as centralised and single-shot, while Chakraa et al.'s survey \cite{chakraa2023} situates the comparison reported in Section VI-C within the broader diversity of centralised, distributed, heuristic, and optimisation-based allocation strategies these reviews identify.

Across the genetic-algorithm allocation literature specifically \cite{eun2009,tan2024,deng2013,yan2024ga}, a consistent pattern emerges: each study reports that population-based search improves assignment quality over simpler heuristics for its own tested problem formulation, whether under heterogeneous-fleet constraints \cite{deng2013}, simultaneous-arrival and resource constraints \cite{yan2024ga}, or other operational variants \cite{tan2024}, but none reports the computation-time cost of that improvement, and none isolates the allocator's own contribution from the trajectory or route-planning component evaluated alongside it. Alqefari and Menai's review of dynamic multi-UAV assignment \cite{alqefari2025} makes the resulting trade-off explicit at the level of method families (market-based \cite{dias2006market}, intelligent-optimisation, clustering-based), identifying computational cost as a property of intelligent-optimisation approaches in general rather than measuring it for a specific allocator; TriSAR's own GA-versus-greedy timing comparison (Section VI-C) supplies exactly this missing measurement for one such allocator.

The collision-avoidance literature builds on Kennedy and Eberhart's introduction of PSO as a population-based search method \cite{kennedy1995}, and separates along a different axis: whether the underlying model reasons over relative position or relative velocity. Position-based reactive methods \cite{khatib1986,fox1997} are computationally cheap but are well known to suffer from local minima, since neither the potential-field formulation \cite{khatib1986} nor the dynamic window approach \cite{fox1997} reasons about a neighbour's velocity, only its instantaneous position. Van den Berg et al.'s reciprocal velocity obstacles \cite{vandenberg2011} were introduced specifically to close this gap by reasoning over relative velocity, at the cost of additional computation. TriSAR's own repulsion term (Section IV-B) sits on the cheaper, position-based side of this divide, and Section VI accordingly reports the local-minima vulnerability this choice inherits directly, rather than assuming it away.

Two hybrid GA-PSO systems most closely resemble TriSAR's own architecture, and both illustrate the same limitation from a different angle. Yan et al. \cite{yan2021pso} embed genetic-style crossover and mutation operators directly inside the PSO population, reporting improvements over \gls{aco} \cite{dorigo2004} and GA baselines across multiple scenario sizes; Xiong and Zhang \cite{xiong2025greapso} share a single grid representation between the GA's encoding and PSO's search space, reporting multi-objective improvements over \gls{nsgaii} and \gls{moeadde}. In both designs the evolutionary and swarm-intelligence contributions are fused into one mechanism rather than kept independently switchable, so neither study can attribute its reported aggregate gain to allocation, trajectory control, or their interaction specifically. TriSAR's architecture deliberately keeps the GA allocator and PSO repulsion term as two independently toggleable components instead (Section IV), so that the factorial ablation in Section V can attribute an observed effect to one component, the other, or their interaction, rather than to the hybrid as a whole.

Read together, the reviewed literature leaves the same gap open from three directions: allocation studies report quality gains without cost measurement or ablation \cite{eun2009,tan2024,deng2013,yan2024ga,alqefari2025}; collision-avoidance studies trade computational cost against local-minima robustness without being evaluated inside a task-allocation pipeline \cite{khatib1986,fox1997,vandenberg2011}; and hybrid GA-PSO systems fuse both coordination layers so tightly that neither can be measured independently \cite{yan2021pso,xiong2025greapso}. TriSAR's controlled $2\times2$ factorial ablation (Section V) is designed to close all three gaps within a single experiment, at the cost of testing only one specific allocator and one specific reactive-avoidance mechanism, not a general claim across the wider method families each belongs to. Against this background, Section III defines the TriSAR system model, including the simulated disaster environment, aerial robot state representation, mission targets, and initial deployment configuration that provide the common experimental basis for the coordination and evaluation methods developed later in the article.

\section{System Model}

\begin{figure}[t]
\centering
\includegraphics[width=0.95\linewidth]{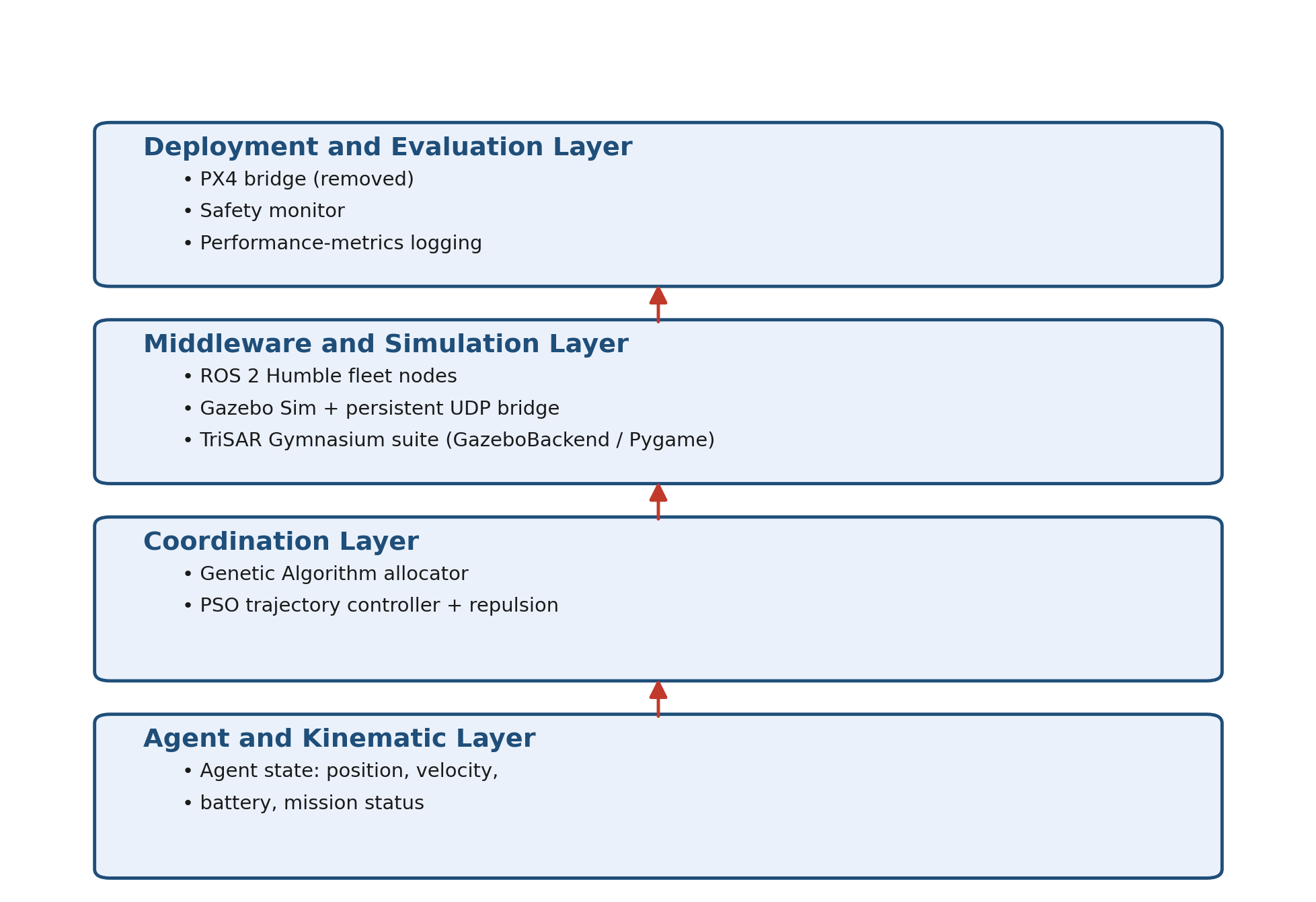}
\caption{TriSAR's four-layer system architecture.}
\label{fig.arch}
\end{figure}

Figure \ref{fig.arch} summarises TriSAR's overall architecture: an agent and kinematic state layer at the base, a coordination layer combining the GA allocator and PSO controller, a middleware and simulation layer bridging to the physics engine, and a deployment and evaluation layer.

\begin{figure}[t]
\centering
\includegraphics[width=0.85\linewidth]{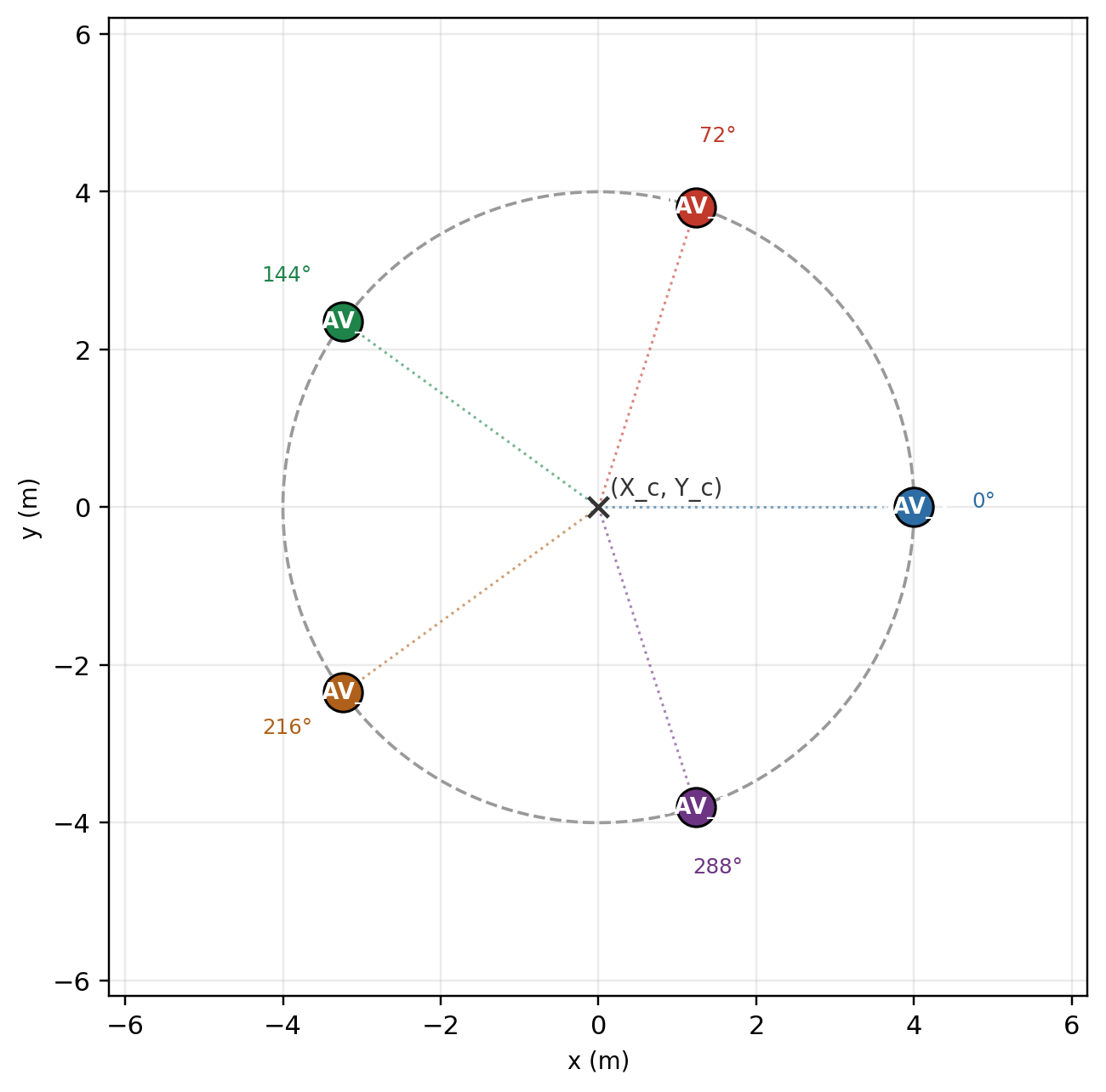}
\caption{Circular ring spawn formation for $N=5$, $R=4.0$ m.}
\label{fig.spawn}
\end{figure}

The environment is a $500\,\text{m}\times500\,\text{m}\times120\,\text{m}$ simulated metropolis containing $N=5$ homogeneous quadcopter agents (3 search-and-rescue agents and 2 dedicated relay agents, the latter assigned fixed rooftop relay stations rather than competing for the $M=8$ targets) and $M=8$ mission targets (4 rooftop, 4 ground-level). Each agent $i$ has position $\mathbf{p}_i\in\mathbb{R}^3$, velocity $\mathbf{v}_i$, maximum speed $v_{\max}=15.0$ m/s, and battery level $b_i\in[0,100]$. Agents are spawned in a circular ring formation of radius $R=4.0$ m,
\begin{equation}
x_i = X_c + R\cos\!\left(\tfrac{2\pi i}{N}\right),\quad y_i = Y_c + R\sin\!\left(\tfrac{2\pi i}{N}\right),
\end{equation}
which provides equal angular dispersion and equal radial spacing around the common spawn centre, avoiding initial positional clustering among UAVs, illustrated in Figure \ref{fig.spawn}.

\section{Proposed Hybrid GA-PSO Coordination}

\begin{figure*}[t]
\centering
\includegraphics[width=0.85\textwidth]{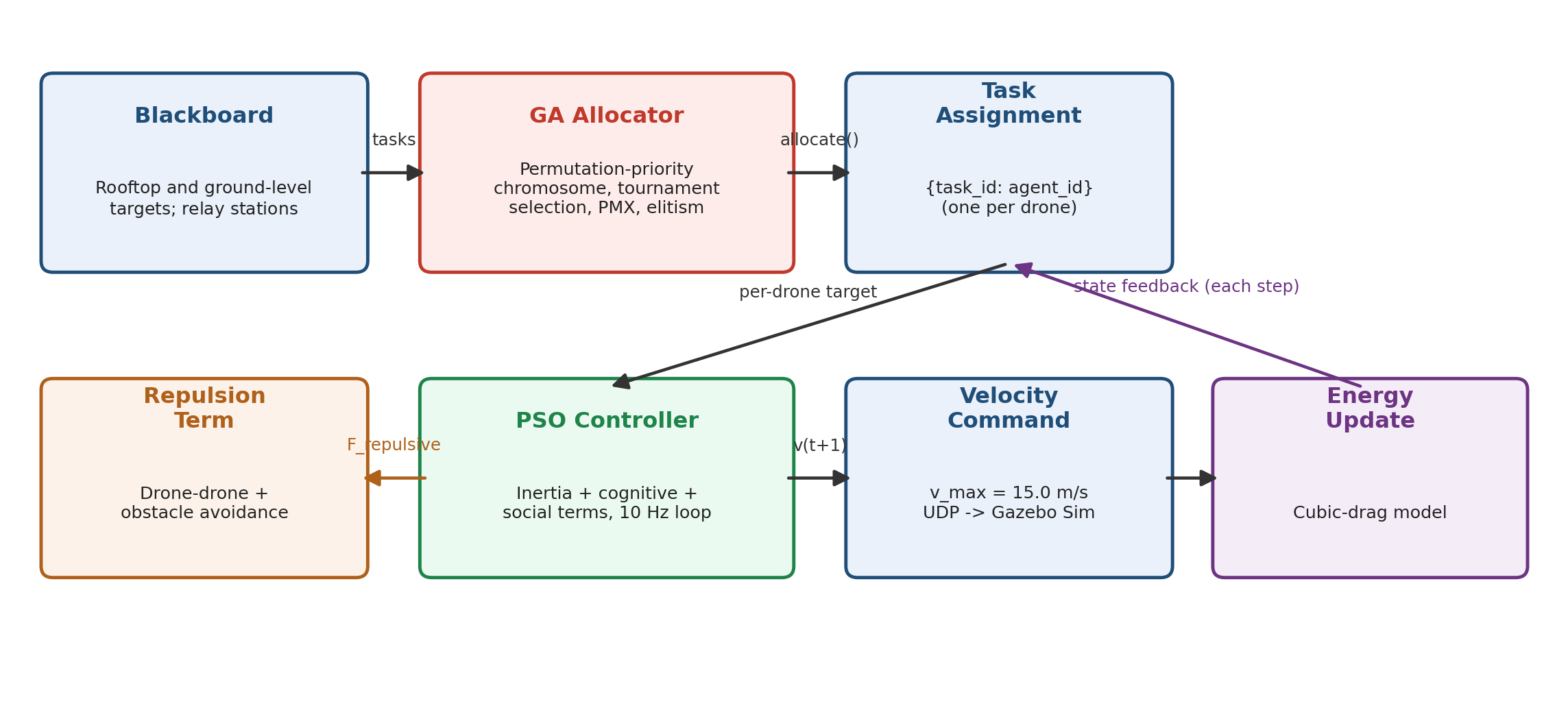}
\caption{TriSAR's GA-PSO coordination pipeline, from task registration through allocation, trajectory control and energy update, with per-step state feedback closing the loop. The blackboard registers the rooftop and ground-level targets defined in Section III; relay agents are assigned fixed rooftop stations separately, rather than competing for these targets.}
\label{fig.pipeline}
\end{figure*}

Figure \ref{fig.pipeline} shows the full coordination pipeline described in this section, from task registration on a shared blackboard through GA-based allocation, PSO-based trajectory control, and the energy update that closes the loop each simulation step.

\subsection{Task Allocation}

A chromosome is a permutation of task indices representing assignment priority order. Decoding walks the chromosome, assigning each task to the agent with the lowest assignment cost $c(\mathrm{a},\mathrm{t})$, evaluated as
\begin{equation}
\begin{aligned}
c(\mathrm{a},\mathrm{t}) = {} & \frac{d(\mathrm{a},\mathrm{t})}{v_{\max}}\left(1+\frac{100-b(\mathrm{a})}{50}\right) \\
& \times\frac{\max(0.1,\,11-u(\mathrm{t}))}{10}\,\tau(\mathrm{type}),
\end{aligned}
\end{equation}
where $d(\mathrm{a},\mathrm{t})$ is the Euclidean distance between agent $\mathrm{a}$ and target $\mathrm{t}$, $b(\mathrm{a})$ is agent $\mathrm{a}$'s current battery level, $u(\mathrm{t})\in[1,10]$ is target $\mathrm{t}$'s urgency (so higher urgency yields lower, more favourable cost), and $\tau(\mathrm{type})=1.25$ for rooftop targets (reflecting additional climb cost) or $1.0$ otherwise. With 3 search-and-rescue agents and 8 targets, decoding proceeds over multiple rounds: after each round assigns one task per available agent, all agents become available again for the next round against remaining tasks, until all 8 are assigned; an unassigned task (not encountered in the tested scenario) would incur a fixed penalty of 500.0 in the chromosome's total cost. Selection uses 3-way tournament; crossover uses \gls{pmx} with probability 0.85; mutation is per-gene swap with probability 0.15; the top-2 individuals are retained via elitism. The population size is 40, with a maximum of 100 generations and early stopping after 15 generations without improvement. Both allocation strategies use the same agent-task cost function above, ensuring that the comparison isolates the effect of the allocation search strategy rather than differences in the underlying cost model: the GA searches over permutations of task priority before decoding the resulting chromosome into agent-task assignments, whereas the greedy baseline performs the corresponding assignment directly without population-based optimisation.

\subsection{Trajectory Control}

Each agent's velocity is updated at 10 Hz following a per-agent local-trajectory variant of the PSO velocity update \cite{kennedy1995}. At each control step, agent $k$ maintains a population of $P=10$ candidate velocity vectors $\mathbf{v}_m$ ($m=1,\dots,P$), freshly re-initialised around the current target-heading direction rather than carried over from the previous step. Each candidate's cost is evaluated at its own projected position, $\mathbf{p}_k+\mathbf{v}_m\Delta t$, so the quantity optimised by the search is the candidate velocity itself, not the agent's physical position. Over $K=5$ inner iterations, indexed by $\tau$ and distinct from the outer 10 Hz control-step index $t$, each candidate is updated as
\begin{equation}
\mathbf{v}_m^{(\tau+1)} = w\mathbf{v}_m^{(\tau)} + c_1 r_1(\mathbf{v}_{\mathrm{best},m}-\mathbf{v}_m^{(\tau)}) + c_2 r_2(\mathbf{v}_{\mathrm{best,g}}-\mathbf{v}_m^{(\tau)}),
\end{equation}
where $\mathbf{v}_{\mathrm{best},m}$ is the lowest-cost velocity candidate slot $m$ has attained across the current control step's iterations, $\mathbf{v}_{\mathrm{best,g}}$ is the lowest-cost candidate found across the whole population at this step, and $w=c_1=c_2=0.5$. Both $\mathbf{v}_{\mathrm{best},m}$ and $\mathbf{v}_{\mathrm{best,g}}$ are tracked per agent across that agent's own candidate population rather than as a single value shared across the fleet, since five agents pursuing distinct assigned targets cannot share one physically meaningful global-best velocity; this is accordingly a per-agent, velocity-space local-trajectory optimiser, not the canonical global-landscape PSO formulation, which searches directly over position. After $K$ iterations, the candidate with the lowest final cost is selected and sent to the drone. Repulsion is applied as a separate adjustment to this selected velocity: for every neighbour or obstacle $j$ within $d_{\mathrm{safe}}=2.5$ m of agent $i$, a unit vector $\hat{\mathbf{r}}_{ij}$ pointing from $j$ towards $i$ is scaled by the fractional violation depth, $(d_{\mathrm{safe}}-d_{ij})/d_{\mathrm{safe}}$, and summed,
\begin{equation}
\mathbf{r}_i = k_{\mathrm{rep}}\!\!\sum_{j:\,d_{ij}<d_{\mathrm{safe}}}\hat{\mathbf{r}}_{ij}\left(\frac{d_{\mathrm{safe}}-d_{ij}}{d_{\mathrm{safe}}}\right),
\end{equation}
with $k_{\mathrm{rep}}=3.0$ m/s. This term is structurally an artificial potential field \cite{khatib1986} and, as discussed in Section VI, inherits its known local-minima susceptibility.

\subsection{Energy Model}

Battery drain follows a cubic-drag model consistent with the parasitic-power term of Zeng et al. \cite{zeng2019},
\begin{equation}
\Delta E = \left(P_{\mathrm{hover}} + C_{\mathrm{drag}}\lVert\mathbf{v}\rVert^{3} + C_{\mathrm{mass}}\lVert\mathbf{a}\rVert\right)\Delta t,
\end{equation}
with $P_{\mathrm{hover}}=0.28$, $C_{\mathrm{drag}}=7\times10^{-5}$, $C_{\mathrm{mass}}=0.05$ (percentage points per second, per corresponding unit). These coefficients are simulation-scale battery-depletion parameters rather than calibrated physical power coefficients; the resulting energy metric is accordingly a normalised simulation proxy used for relative comparison between experimental conditions, not for prediction of real UAV energy consumption.

\section{Experimental Setup}

All results use a fixed scenario (5 drones, 8 targets: 4 rooftop, 4 ground-level) executed in a Gazebo Sim (Harmonic) backend bridged to a Python control loop via a persistent \gls{udp} socket. Four variants were each run for $n=30$ independent episodes: \textbf{Full} (GA + PSO with repulsion); \textbf{No-GA} (greedy nearest-cost allocator + PSO with repulsion); \textbf{No-Repulsion} (GA + PSO, repulsion disabled); \textbf{Floor} (greedy allocator, repulsion disabled). Each episode uses an independently drawn random seed; ground-level target coordinates are additionally perturbed each episode by uniform jitter of $\pm2$ m on each axis, while rooftop coordinates, agent count, and environmental conditions are held fixed across episodes and are not separately matched between the four variants. Reported metrics are mission completion, execution steps, path length, cumulative fleet energy consumed, and collision-threshold violations, given as mean $\pm$ standard deviation. A target is considered reached, and is removed from an agent's task queue, when the agent comes within 5.0 m of it in three-dimensional Euclidean distance or within 4.0 m in horizontal (two-dimensional) distance, whichever condition triggers first; arrival is evaluated instantaneously, with no dwell-time or additional action required. Mission completion requires every target to be reached by some agent.

A collision-threshold violation is operationally defined as a pairwise Euclidean separation between two agents falling below $d_{\mathrm{safe}}=2.5$ m, evaluated once per control step on realised (not projected) positions; a violation between an agent and a restricted building zone is defined analogously using that zone's radius plus a fixed clearance. This threshold is a distance-based safety criterion rather than a physical Gazebo contact event, and we refer to violations rather than collisions accordingly where precision matters. A single sustained violation spanning several consecutive control steps is recorded as multiple discrete violation-steps rather than collapsed into one incident; this convention is applied uniformly across all four conditions. Drone-drone and drone-obstacle violations are logged separately but combined into a single count for the statistics reported below, since the ablation design tests the repulsion mechanism as a whole rather than its inter-agent and obstacle-avoidance components individually; no drone-obstacle violations were recorded in the reported batch.

\section{Results and Discussion}

All results reported in this section are drawn from the single, consistent $n=30$-per-condition ablation batch described in Section V; no results from separately-run batches are combined with these figures, so that all comparisons in Table \ref{tab.ablation} are directly comparable.

\subsection{Statistical Analysis}

We analyse the $2\times2$ factorial design directly: allocator (GA vs. greedy) and repulsion (enabled vs. disabled) as two factors, each with $n=30$ episodes per cell. For the three continuous efficiency metrics (steps, path length, energy), we report Welch's unequal-variances $t$-tests computed directly on the per-episode data, with 95\% \glspl{ci} and Hedges' $g$, since GA-allocated conditions show substantially higher variance than greedy-allocated conditions (Table \ref{tab.ablation}). Table \ref{tab.anova} reports the result: with repulsion enabled, no GA-versus-greedy comparison's CI excludes zero; with repulsion disabled, every comparison's CI excludes zero, with $|g|$ between 0.92 and 1.76. This pattern is consistent with an allocator-repulsion interaction, though no formal interaction term was estimated; it directly answers RQ1 conditionally rather than uniformly: GA shows no detectable efficiency advantage when repulsion is enabled, but a significant advantage on all three metrics when repulsion is disabled.

For the collision-threshold-violation outcome, Full and No-GA recorded zero violation-steps across all 30 episodes; No-Repulsion and Floor recorded means of $2.70\pm3.22$ and $3.03\pm1.96$ respectively. A Mann-Whitney $U$ test comparing Floor against No-GA is significant ($U=885$, $p=4.03\times10^{-12}$, rank-biserial $r=0.97$), confirming repulsion's protective effect under greedy allocation specifically. The equivalent test under GA allocation (Full vs.\ No-Repulsion) is also significant ($U=675$, $p=1.26\times10^{-5}$, rank-biserial $r=0.50$), confirming repulsion's protective effect under GA allocation as well. This answers RQ2 with strong statistical support under both allocators. For RQ3, the two effect sizes differ substantially in magnitude ($r=0.97$ under greedy versus $r=0.50$ under GA), a pattern directionally consistent with the allocator dependence observed for the efficiency metrics above; as with RQ1, however, no formal interaction term was estimated for the collision outcome either, so this pattern is reported as suggestive rather than as a formally confirmed interaction.

\begin{table}[t]
\centering
\caption{Welch's $t$-tests, GA vs.\ greedy, with 95\% CIs and Hedges' $g$.}
\label{tab.anova}
\small
\begin{tabular}{@{}llc@{}}
\hline
Metric & Comparison & Mean diff.\ [95\% CI], $g$ \\
\hline
Steps  & Repulsion ON  & 5.30~[$-7.70$, 18.30], $g{=}0.21$ \\
       & Repulsion OFF & $-7.67$~[$-12.00$, $-3.34$], $g{=}{-}0.92$ \\
Path   & Repulsion ON  & $-4.54$~[$-14.58$, 5.50], $g{=}{-}0.24$ \\
       & Repulsion OFF & $-16.42$~[$-21.25$, $-11.59$], $g{=}{-}1.76$ \\
Energy & Repulsion ON  & 3.99~[$-2.44$, 10.42], $g{=}0.32$ \\
       & Repulsion OFF & $-2.99$~[$-4.40$, $-1.58$], $g{=}{-}1.09$ \\
\hline
\end{tabular}
\end{table}

\subsection{Ablation Study}

\begin{figure*}[t]
\centering
\includegraphics[width=0.8\textwidth]{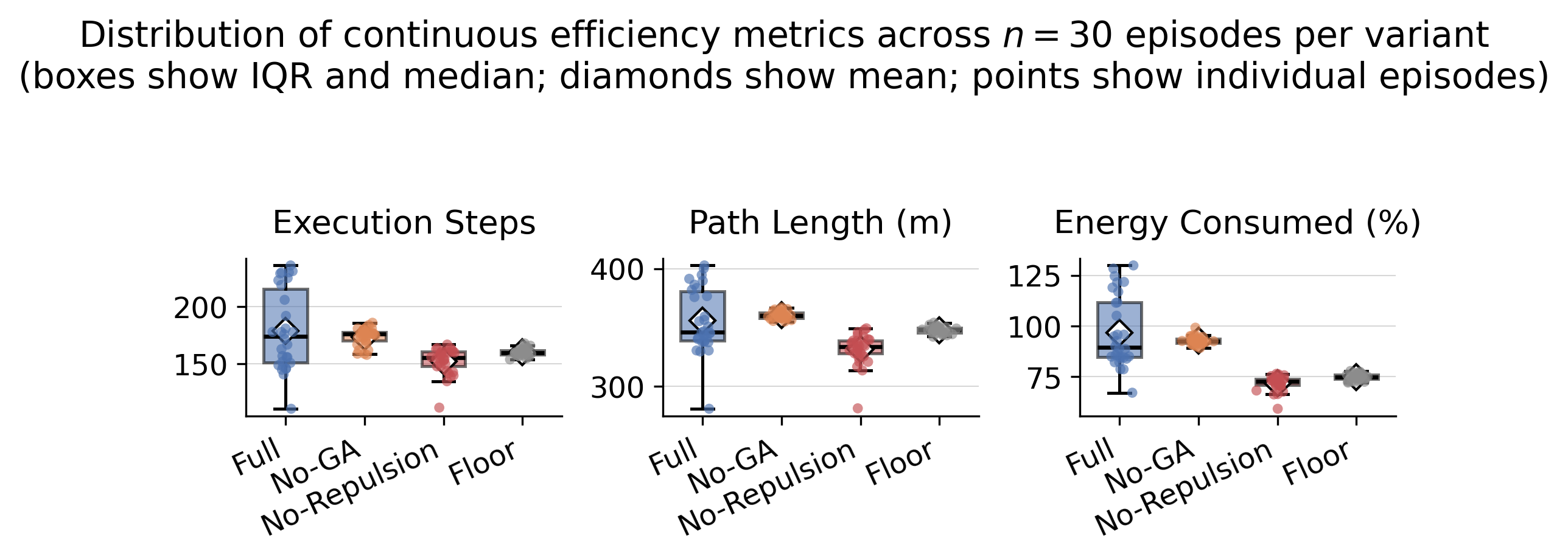}
\caption{Distribution of the three continuous efficiency metrics across all $n=30$ individual episodes per variant. Boxes show the interquartile range and median; diamonds show the mean; points show every individual episode.}
\label{fig.ablation}
\end{figure*}

Table \ref{tab.ablation} and Figure \ref{fig.ablation} report the four-variant comparison underlying the statistical analysis above.

\begin{table}[t]
\centering
\caption{Four-variant ablation comparison ($n=30$ per variant).}
\label{tab.ablation}
\scriptsize
\setlength{\tabcolsep}{2.5pt}
\begin{tabular}{@{}lcccc@{}}
\hline
Variant & Steps & Path (m) & Energy \% & Viol. \\
\hline
Full & $179.07\pm34.15$ & $355.85\pm26.74$ & $96.85\pm17.12$ & $0.00\pm0.00$ \\
No-GA & $173.77\pm7.46$ & $360.39\pm3.14$ & $92.86\pm1.94$ & $0.00\pm0.00$ \\
No-Repulsion & $152.63\pm11.20$ & $331.35\pm12.70$ & $71.81\pm3.50$ & $2.70\pm3.22$ \\
Floor & $160.30\pm3.23$ & $347.77\pm2.76$ & $74.80\pm1.55$ & $3.03\pm1.96$ \\
\hline
\end{tabular}

\vspace{0.15em}
\footnotesize{Episodes with $\geq$1 violation, $n$(\%): Full 0/30 (0\%); No-GA 0/30 (0\%); No-Repulsion 15/30 (50.0\%); Floor 29/30 (96.7\%).}
\end{table}

Figure \ref{fig.ablation} shows the full per-episode distribution underlying Table \ref{tab.ablation} rather than only the mean and standard deviation. All four distributions are visibly right-skewed, with the mean sitting above the median in every case, and each variant contains at least one high-value outlier episode. This skew does not change the conclusions above, which use Welch's test specifically because it does not assume equal variance, but it is a more complete picture of the underlying variability than the summary statistics alone convey, and is reported here directly rather than obscured behind symmetric error bars.

\textbf{Repulsion reduces collision-threshold violations, confirmed under both allocators.} Floor (greedy, repulsion disabled) recorded a mean of $3.03\pm1.96$ violation-steps per episode against zero for No-GA (greedy, repulsion enabled); this difference is statistically confirmed (Mann-Whitney $U=885$, $p=4.03\times10^{-12}$, rank-biserial $r=0.97$). No-Repulsion (GA, repulsion disabled) recorded a mean of $2.70\pm3.22$ violation-steps against zero for Full (GA, repulsion enabled); this difference is also statistically confirmed (Mann-Whitney $U=675$, $p=1.26\times10^{-5}$, rank-biserial $r=0.50$), though with a smaller effect size than under greedy allocation. At the episode level, 29 of 30 Floor episodes and 15 of 30 No-Repulsion episodes recorded at least one violation-step, against zero for both repulsion-enabled conditions.

\begin{figure*}[t]
\centering
\includegraphics[width=0.9\textwidth]{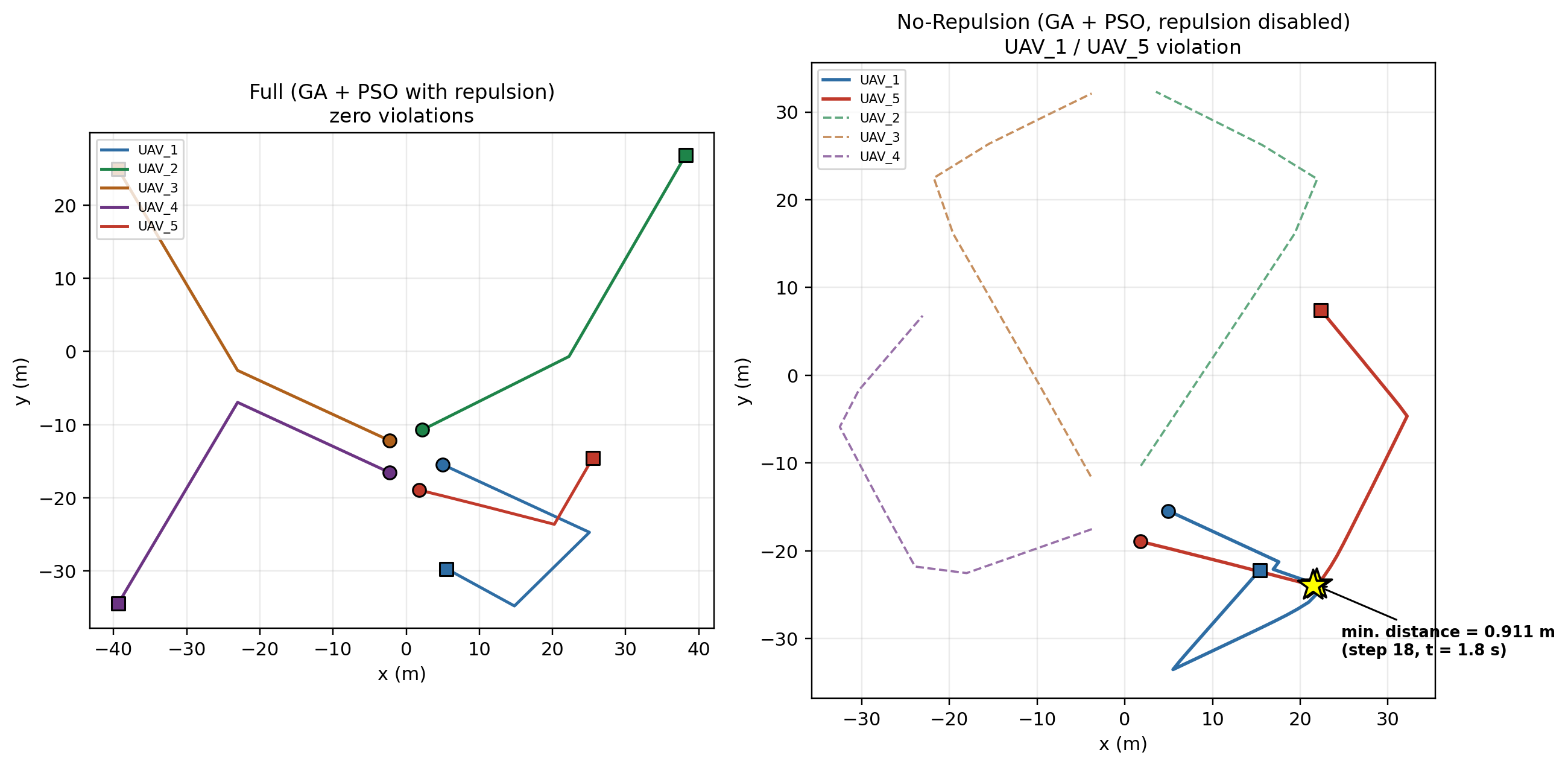}
\caption{Representative flight trajectories (top-down, $x$-$y$ plane) for a violation-free Full episode (left) and a violation episode under No-Repulsion (right). Circles mark starting positions, squares mark final positions, and stars mark the verified point of closest approach between UAV\_1 and UAV\_5 (0.911~m at step 18) in the violation episode.}
\label{fig.traj}
\end{figure*}

Figure \ref{fig.traj} shows representative flight paths illustrating the violation mechanism (generated under an earlier system version, not the batch reported in Table \ref{tab.ablation}). In the illustrated No-Repulsion episode, two UAVs converge to a minimum separation of 0.911~m, below the 2.5~m collision-threshold defined in Section V, before separating again. The corresponding Full episode shows all five drones maintaining separation throughout.

\textbf{The GA allocator shows an efficiency advantage only when repulsion is disabled.} Contrary to a uniform directional expectation \cite{tan2024}, the Full variant is the worst of the four on steps and energy consumption and second-worst on path length (behind No-GA); Table \ref{tab.anova} shows none of these repulsion-enabled differences reach significance. With repulsion disabled, GA (No-Repulsion) is the best of the four on all three metrics, and every GA-vs-greedy comparison in that condition is significant with large effect sizes (Table \ref{tab.anova}). A plausible mechanism, offered as interpretation, is that active repulsion's reactive detours dominate route shaping for both allocators alike, masking whatever advantage GA's task ordering otherwise provides. We report this as a scenario-scale-dependent, repulsion-conditional finding: Yan et al.\ \cite{yan2021pso} and Xiong and Zhang \cite{xiong2025greapso} both report GA/hybrid advantages that grow with problem size and asymmetry, and a natural extension of this work is to test whether TriSAR's own GA allocator shows an analogous scale-dependent advantage at larger $N$ and $M$ or under asymmetric target distributions.

\subsection{Computational Cost}

Allocation wall-clock time was not captured for the main $n=30$ canonical batch; we added this logging afterwards and report it here from a separate $n=5$ verification sample per allocator, run under otherwise identical conditions on the same development machine (AMD Ryzen 5 4600H, 3.00 GHz, 8 GB RAM) used for all reported episodes. GA allocation time clustered tightly (mean $86.53\pm13.83$ ms, range 67.6--109.3 ms), with early stopping firing consistently after 16--19 generations in every run, well short of the 100-generation cap, indicating this is representative behaviour rather than an artefact of a single lucky or unlucky run. Greedy allocation was both faster and more consistent (mean $0.174\pm0.039$ ms, range 0.11--0.21 ms), giving an approximate $500\times$ computational overhead for GA relative to greedy allocation. In absolute terms this overhead is small relative to per-episode wall-clock flight time (multiple seconds), so it does not materially affect the mission-efficiency metrics reported above; it is, however, a real and consistent cost that any future scale-up of the GA allocator should account for, since the 100-generation ceiling was never approached at this scenario size and larger, more complex assignment problems may behave differently.

\subsection{Limitations}

Five limitations bound the interpretation of these results. First, the ablation was conducted at a single, small, symmetric scenario scale; the finding regarding GA performance should not be generalised beyond it without further testing at larger scale. Second, no comparison against established external baselines (Ant Colony Optimisation \cite{dorigo2004}, market-based allocation \cite{dias2006market}, or reciprocal velocity obstacles) was conducted; the ablation isolates TriSAR's own components against internal simplifications of themselves, not against competing published methods. Third, the computational-cost measurement above is drawn from a separate $n=5$ verification sample per allocator rather than the full $n=30$ canonical batch, since allocation timing was not logged when that batch was originally collected; the tight clustering observed gives reasonable confidence in the reported figures, but a full $n=30$ timing sample was not collected before submission. Fourth, all results are obtained in physics-based simulation (Gazebo Sim) using perfect simulator ground truth; sensing and communication effects were not modelled at all, rather than modelled in simplified form, and no simulation-to-real transfer has been attempted or evaluated. Fifth, the 2.5 m collision threshold is a distance-based safety criterion evaluated on simulated agent positions, not a physical Gazebo contact event; the reported violation counts should be read accordingly and would need re-derivation against a specific airframe's physical dimensions before informing a real deployment.

\section{Conclusion and Future Work}

We presented TriSAR, a hybrid GA--PSO coordination architecture evaluated for multi-UAV disaster response, and reported a controlled four-variant factorial ablation study rather than aggregate self-reported performance alone. The results provide statistically-supported evidence that local reactive collision avoidance measurably reduces collision-threshold violations under both allocation strategies in this scenario (Mann-Whitney $U=885$, $p=4.03\times10^{-12}$ under greedy allocation; $U=675$, $p=1.26\times10^{-5}$ under GA allocation), with a substantially larger effect size under greedy allocation, and a conditional finding regarding the GA allocator's efficiency: no detectable advantage over greedy when repulsion is enabled, but a significant advantage on all three efficiency metrics when repulsion is disabled (Welch's $t$-tests, Table \ref{tab.anova}). Future work should extend the ablation to larger, asymmetric scenarios to test for scale-dependent GA advantage, fit a formal interaction model to confirm the allocator-repulsion pattern reported here, benchmark against reciprocal velocity obstacles to directly address the local-minima vulnerability inherited from the potential-field-like repulsion term, compare against established external baselines (ACO \cite{dorigo2004}, market-based allocation \cite{dias2006market}) to situate TriSAR's performance within the wider literature rather than only against internal ablations of itself, extend the fleet to heterogeneous robot capabilities rather than homogeneous quadcopters, and pursue simulation-to-real validation of the coordination architecture on physical hardware.

\section*{Reproducibility}

Source code, simulation configurations, experiment scripts and data-processing pipelines required to reproduce the reported results are publicly available at \url{https://github.com/Aditya-1711/TriSAR}.

\bibliographystyle{IEEEtran}
\bibliography{paper}

\end{document}